\documentclass{article}

\usepackage[preprint]{corl_2026}
\usepackage{float}
\usepackage{graphicx}
\usepackage{url}
\usepackage{booktabs}
\usepackage{xcolor}

\title{How Should a Home Robot Update Its Memories When Human Routines Change?}

\author{
  Oliver Wang \qquad Mani Srivastava\\
  University of California, Los Angeles\\
  \texttt{owang22@ucla.edu}
}

\begin{document}
\hypersetup{pdftitle={How Should a Home Robot Update Its Memories When Human Routines Change?},pdfauthor={Oliver Wang, Mani Srivastava},pdfsubject={},pdfkeywords={}}
\maketitle
\raggedbottom

\begin{abstract}
A home robot that helps with chores needs to know where objects are, but people move objects according to routines that change and later change back. The robot's memory must adapt to the new routine without erasing experience that may become useful again. One approach is to have a language model maintain written summaries of the home from the robot's own observations, but existing evaluations of such memories use static worlds. We explore simulated households where a resident gets sick, recovers, and later gets sick again. We compare a rule that keeps every room an object has been seen in, language-model memories that store only written summaries, and language-model memories that can also read the raw history of sightings. The rule outperforms every summary-only memory, and giving language models the sighting history closes the gap. Surprisingly, writing summaries on top of the history adds nothing we can measure. When the illness returns, methods that keep the history search better than they did the first time, while summary-only methods show no consistent improvement. Deleting the first illness from memory removes all of the rule's gain and part of theirs. The summaries recorded where objects went during the illness but rarely distinguished sick-day patterns from normal routines. Our findings suggest that common approaches to summarizing robot memories struggle to remember facts about a routine that might return, highlighting the need for future benchmarks to evaluate this aspect of long-term robot deployment.
\end{abstract}

\section{Introduction}
\label{sec:intro}

Robots deployed in homes must keep learning about the specific home they work in. A robot that helps with chores should know where objects are, and where objects are depends on the people who live there and on what those people are doing. If someone gets sick and stays home for a week, maybe their laptop that was usually stored in a backpack now tends to be open all day on their bedroom desk. When they recover, the laptop goes back to the backpack. A robot's memory has to do three things at once. It must adapt during the sick period, recognize when to switch back to the healthy routine, and remember the sick-week experience (so that the next time this person is sick the robot already knows where to look).

One approach is to give the robot a language model that maintains written summaries of the home, revising them each night from what the robot saw and reading them each day when deciding where to search~\citep{park2023generative,packer2023memgpt,zhang2026ace,pi2026mem}. A language model could notice that several objects moved for one shared reason, or recognize a situation it has seen before. However, Zhang et al.~\citep{zhang2026faulty} show that when a language model repeatedly rewrites its own memory, the memory accumulates errors, and that simply keeping the raw episodes is competitive with every summarization method they test. Their worlds never change. Robot memory benchmarks test recall within one episode or as unrelated sessions accumulate~\citep{dai2026robomme,lei2026robomemarena,robomme2026interference}, and scene memories keep a map current as objects move~\citep{liu2024dynamem,anwar2025remembr,schmid2024khronos}. None of these test whether a memory written over weeks keeps what one routine taught it for the day that routine returns.

We test this in simulated households where one resident gets sick, recovers, and later gets sick again. The robot answers object-location questions by observing up to three rooms, learns only from what it sees, and is never told that anyone is sick. We compare a simple rule that keeps every room an object has been seen in, language-model memories that store only written summaries, and language-model memories that can also read the raw history of sightings. Two conclusions come out of this. First, what helped the robot search was keeping its past sightings available. The summaries carried less than the sighting history, and a model that read the history did no better for also writing summaries. Second, the summaries adapted to the illness but did not retain it: they recorded where objects went and never that anyone was sick. When the illness returned, the methods that kept the history did better, mostly by reusing stored sightings. We argue that written robot memories should also be evaluated on what they keep for a routine that returns.

\begin{figure}[htb]
    \centering
    \includegraphics[width=0.55\linewidth]{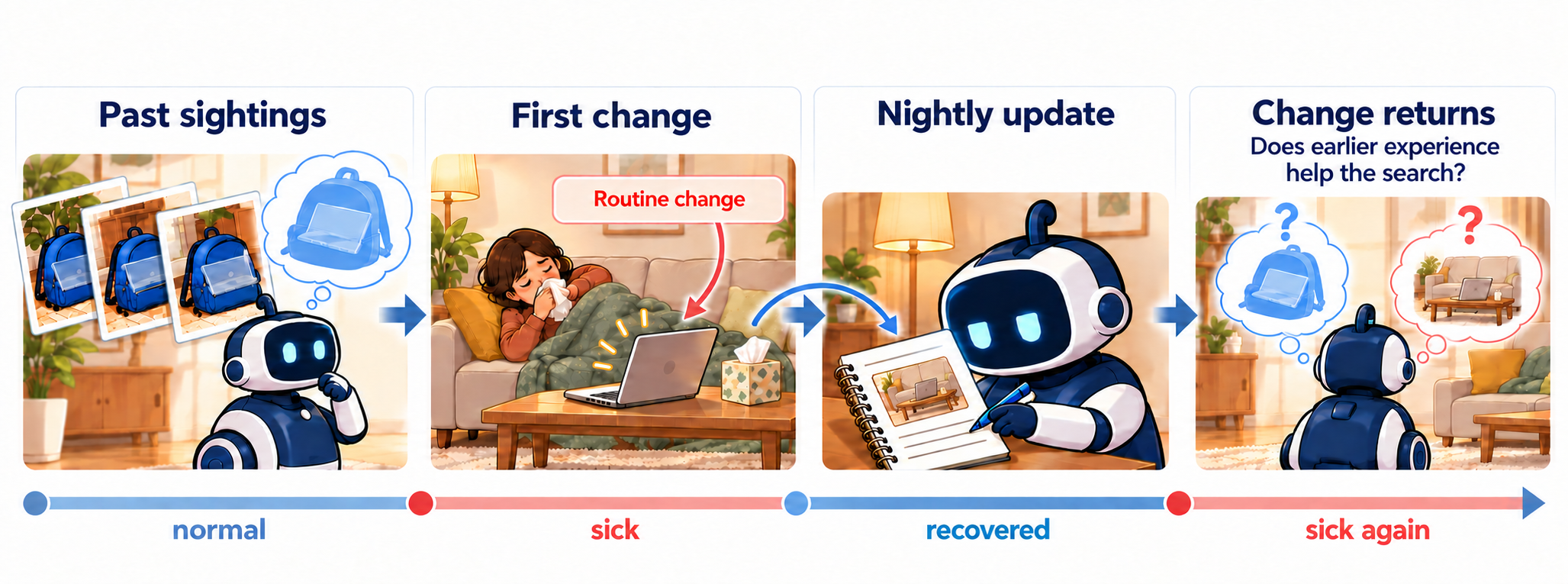}
    \caption{\textbf{Learning from searches across a recurring change.} The robot gathers occasional sightings about the home, the objects, and the resident. When the resident is sick, the robot may not know that immediately but can see the change in daily living patterns. The resident eventually recovers, but they may get sick again in the future.}
    \label{fig:overview}
\end{figure}

\section{Setting and Methods}
\label{sec:setting}

\paragraph{The search-and-update loop.}
A question arrives: where is a particular object? The robot observes a room, and up to two more if the object is not there. Each observation records every object in that room with its spot and the time; we call each record a \emph{sighting}, and an object's accumulated sightings its \emph{sighting history}. Each observation also records which residents are in the room, by name. Some methods also have a language model write text about the household; we call that text its \emph{summaries}, each entry a \emph{note}, and the step where the model revises its summaries each night the \emph{nightly update}. The nightly update sees the day's sightings, including the resident names, and the current summaries. Nothing ever tells any method that a resident is sick. We score \emph{first room right} and \emph{found within three rooms}.

\paragraph{The hidden change in routine.}
Each simulated household has two or three residents, eight or nine rooms with about forty spots between them, and 76 to 104 objects that residents move as they go about their day. Days 1--13 follow the normal routine. On days 14--23 one resident is sick and stays home, which changes where their objects tend to be: a glass becomes more likely to be on the bedroom nightstand than in the kitchen. Days 24--31 return to normal. The 50-day run repeats the illness on the same resident on days 32--41 and returns to normal on days 42--49. The second illness is a fresh draw from the same distribution, so its days differ from those of the first. We call an object a \emph{moved object} if its most common daytime spot differs between normal and sick days, and a \emph{room-changing object} if it also changes room; Section~\ref{sec:second} uses the latter, since a move within one room does not affect where the robot searches. Within a household, every method runs on the same frozen world and question schedule and differs only in which rooms it observes. We use three runs and never pool them: ten households at 8 questions a day for 31 days; three households at 24 a day for 50 days, with the second illness; and five households at 4, 8 and 24 a day. Appendix~\ref{app:setting} gives the statistics.

\paragraph{Methods.}
\emph{Last-seen} uses no language model. For each object it keeps every room the object has been seen in, newest first, observes them in that order, and updates the list after every room it opens. All other methods use \texttt{Qwen3.8-27B}~\citep{qwen38modelcard} at temperature zero. The \emph{summary-only memories} never show the model an object's sighting history. The \emph{claim store} is a list of notes, each with a condition under which the claim is speculated to hold true. The \emph{ACE-style playbook} adds, rewrites and merges notes through nightly reflection~\citep{zhang2026ace}, and is our variant of the published work (Appendix~\ref{app:ace}). \emph{History and notes} writes notes in the same format with the same edits, but its prompts ask about the people in the home and cap the notes at 16, and its room choice also sees the asked object's sighting history. \emph{History only} uses that room-choice prompt with the notes removed and never writes summaries, so it differs from history and notes only in the notes. Code and prompts: \url{https://github.com/nesl/memory-for-home-robots}.

\section{Results}
\label{sec:results}

\subsection{Summaries Carry Information, But Less Than the Sighting History}
\label{sec:rule}

\begin{table}[htb]
  \centering
  \small
  \begin{tabular}{lcc}
    \toprule
    Method & First room right (\%) & Found within 3 rooms (\%) \\
    \midrule
    History and notes             & 85.6 & 96.6 \\
    History only (no notes)       & 85.2 & 95.9 \\
    Last-seen (no language model) & 82.1 & 95.0 \\
    Claim store                   & 77.0 & 86.9 \\
    ACE-style playbook            & 76.0 & 85.3 \\
    \bottomrule
  \end{tabular}
  \caption{\textbf{Accuracy over all questions, ten households, 8 questions a day}, 2,480 questions per method. Last-seen beats both summary-only memories. Last-seen, history only and history and notes are not reliably different from one another.}
  \label{tab:accuracy}
\end{table}

Do the summaries help at all? Table~\ref{tab:accuracy} says yes, but less than the sighting history does. A claim store that writes notes but cannot read them scores about 20 points lower, so the notes carry information. Yet last-seen, which stores only rooms in recency order, beats every summary-only memory in every household. Part of its lead comes from updating after every room it opens: held to the summaries' nightly schedule, it is level with the claim store.

Giving the language model the sighting history closes the gap. History only, which never writes a summary, scores within 1.6 points of history and notes in all three runs, inside our noise floor, so the notes add nothing measurable on top of the history. Adding only the history to the claim store's room-choice prompt raises a rerun of it from 75.8 to 84.1 (Appendix~\ref{app:log}). What does the history give the robot that the summaries do not?

\subsection{Keeping the Sighting History Pays Off When the Illness Returns}
\label{sec:second}

\begin{figure}[htb]
    \centering
    \includegraphics[width=0.62\linewidth]{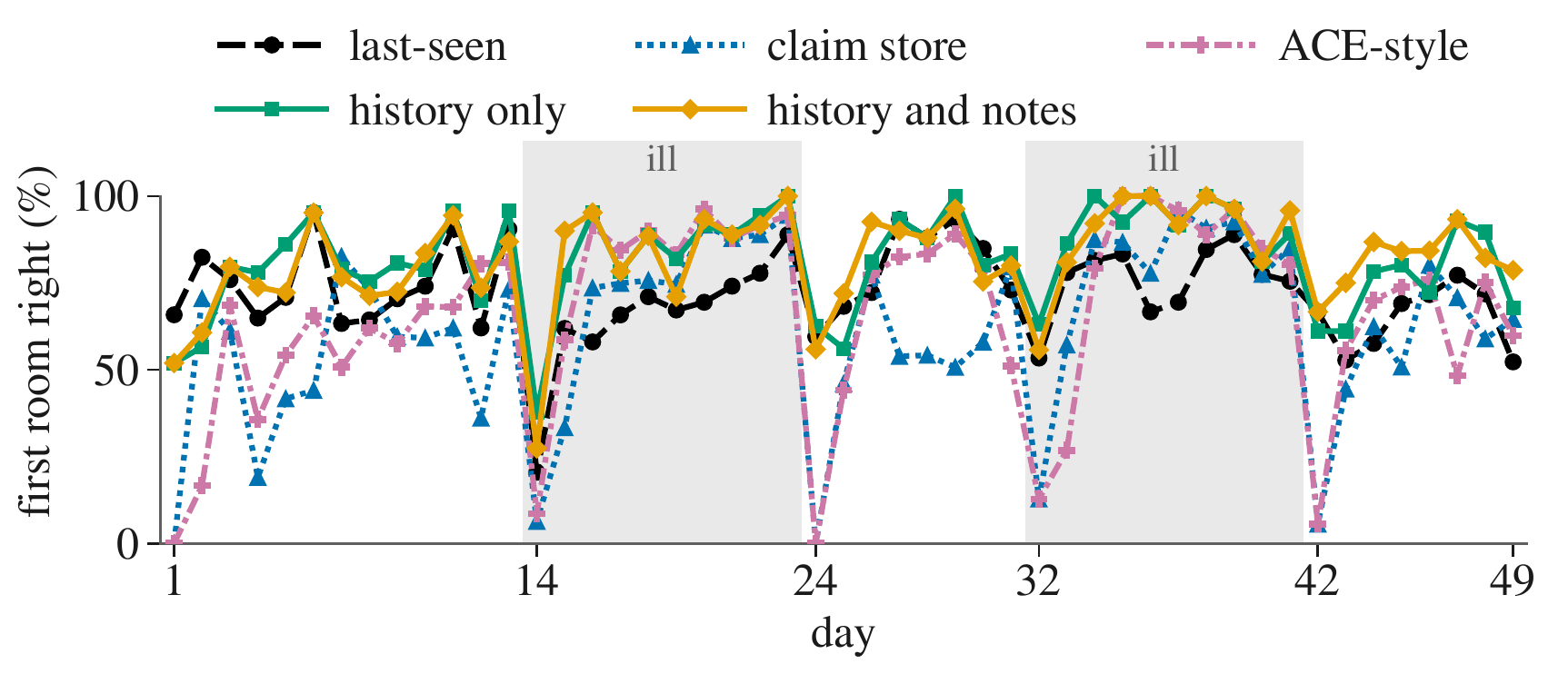}
    \caption{\textbf{The same illness, twice.} First room right by day on the 14 room-changing objects, three households at 24 questions a day; sick days 14--23 and 32--41 shaded.}
    \label{fig:second}
\end{figure}

If a memory retained experience from the first illness, it should drop less on day 32 than on day 14. Measured against each method's mean over the four preceding days, last-seen and the two history methods drop about half as far the second time, while the claim store and the ACE-style playbook drop about as far both times (Figure~\ref{fig:second}). With three households and wide two-SE bands (Appendix~\ref{app:setting}), we read this as the direction of an effect.

\begin{figure}[htb]
    \centering
    \includegraphics[width=0.6\linewidth]{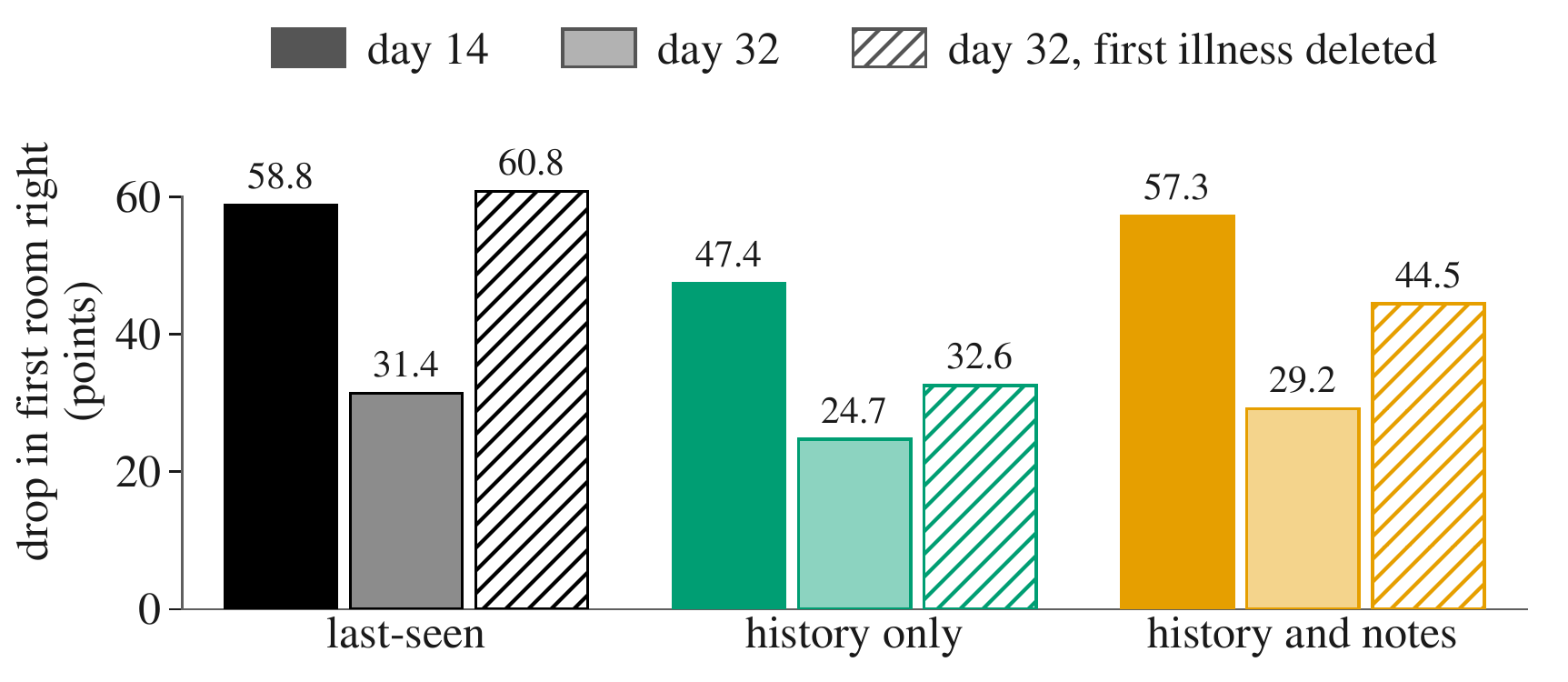}
    \caption{\textbf{Deleting the first illness.} Drop in first room right on the 14 room-changing objects. Hatched bars replay the method with every sighting from days 14--23 deleted as it arrived.}
    \label{fig:delete}
\end{figure}

Does the gain come from stored sightings? We replayed last-seen and the two history methods with every sighting from days 14--23 deleted as it arrived (Figure~\ref{fig:delete}). Last-seen's day-32 drop returns to its day-14 size, so its gain comes from the first illness in its room list. The history methods lose part of their gain. Because the robot often sees a moved object in its new room within hours, our test separates retention from fast relearning only in part (Appendix~\ref{app:delete}).

\subsection{Why Did the Summaries Lose What the Sighting History Kept?}
\label{sec:why}

On night 31 of the ten-household run, a week after recovery, the three summary-writing methods together held 426 notes that place an object in a room it moved to during the illness. So the summaries recorded where things went. Most of these notes attach a time of day or ``always.'' Twelve name a person being home, which is also true on well days, and the rest attach other conditions. None mentions the resident being sick (Appendix~\ref{app:conditions}). One note followed the glass to the bedroom and back, then dropped the bedroom spot (Figure~\ref{fig:life}). When we told history and notes the cause on night 13, it predicted where the resident's objects would go, but in seven of eight runs a later update dropped the cause (Appendix~\ref{app:told}).

\section{Discussion}
\label{sec:discussion}

Every method adapted within the first illness, but only the methods that kept the sighting history did better when the illness came back, and deleting the first illness removed all of the rule's gain and part of theirs. Zhang et al.~\citep{zhang2026faulty} found raw episodes competitive with language-model memories in static worlds; a change that returns is what separates adaptation from retention. Robot memory benchmarks~\citep{dai2026robomme,lei2026robomemarena,robomme2026interference} could test retention with sparser observation, a tighter search budget, and routines whose rooms overlap so that only the cause tells them apart. Keeping the raw sightings costs under 75 KB a day, and Appendix~\ref{app:limits} lists our limitations.

Our results also highlight one way written summaries fail. A summary written during the illness did not know that sickness was a variable worth keeping; it saw a glass on a nightstand in the afternoon and kept the time. A language model has the common sense to guess that a glass by the bed means someone is unwell, and ours did exactly that when we told it, but it later wrote the cause out of its notes and kept the glass. A memory that keeps its sightings can fall back on them when the routine returns, while one that only summarizes has to decide what to keep before it knows what will matter. Whether a language model can guess the cause on its own, keep it, and retrieve it when the routine returns is the open question this setting raises.

\bibliography{references,references_additions}

\appendix

\section{Setting Details and Statistics}
\label{app:setting}

\paragraph{Households.}
Day 0 is a night with no questions. Between 10 and 20 of the objects in each household are asked about. Questions cover eleven kinds of object, including books, mugs, glasses, water bottles, chargers and tablets. The question schedule depends only on the question bank and is identical for every method. The sets of moved and room-changing objects are read from the simulator's ground truth and are the same for every method.

\paragraph{Simulator dynamics.}
Each household has a hidden per-day situation that selects which activities the residents do. A day is realised as a sequence of activity bouts. Every object placement passes through one decision function in fixed order: an event rule if one applies, then a tidy pass, then placement on arriving home, then placement during a home activity, then a fallback if the intended spot is blocked, then a random whim. An object stays where it was put until a resident moves it. Across the households 76 to 104 objects exist, 52 to 74 ever move, and 106 to 213 moves happen a day. The illness event removes 35 activities from the sick resident's day, adds nine bouts drawn from eight distinct activities, sets the resident's energy parameter lower, and adds eight placement rules. Those eight rules are the whole of what moves the asked-about objects. Two kinds of object serve as controls: classes with no illness rule (razor, towel, notebook) and ruled classes owned by a resident who is never sick. The second illness reuses the same event but redraws the day. Of the 22 moved objects of the first illness, 8 do not change room in the second, which shows that the second illness is a fresh draw. The world is frozen in a household's question bank before any method runs.

\paragraph{Observation budget.}
The number of questions per day sets how many rooms the robot observes. For the methods reported here, at 8 a day it observes 8.9--10.9 rooms, at 24 a day 26.5--31.0. Methods that search worse observe more.

\paragraph{Statistics.}
We compare methods by paired differences within each household, averaged across households, and report two standard errors (two SE) of that average. Generation at temperature zero is not exactly repeatable on our server, so we measured a noise floor from two runs whose prompts are byte-identical over days 0--13 of a separate seven-household run. The mean absolute difference was 1.9 points on first room right and 2.2 on found within three rooms, with a largest single-household difference of 6.3 and 7.3. We call a difference clear when it exceeds both the noise floor and two SE with the same sign in every household. Questions within a household are not independent, so the household is the unit of comparison.

\paragraph{Table 1 margins.} Last-seen leads the claim store by $+5.1$ / $+8.2$ points and the ACE-style playbook by $+6.0$ / $+9.7$. Hiding the claim store's notes costs $23.5$ / $21.1$. History and notes leads last-seen by $+3.5$ / $+1.5$ and history only by $+3.1$ / $+0.8$; neither clears two SE or has the same sign in every household. History only trails history and notes by $0.4$ / $0.7$, inside the noise floor, with the sign split 5 of 10. History only and history and notes lead the claim store by $+8.2$ / $+9.0$ and $+8.6$ / $+9.7$, with the same sign in 9 and 10 of 10 households.

\paragraph{Last-seen updated nightly.} This variant keeps the same room list as last-seen but adds the day's sightings to it only at night, on the summaries' schedule. On first room right it trails last-seen by $5.2 \pm 4.0$ points, lower in 8 of 10 households, and is level with the claim store ($-0.2 \pm 6.5$).

\begin{table}[H]
  \centering
  \small
  \setlength{\tabcolsep}{4pt}
  \begin{tabular}{lccccc}
    \toprule
    Method & Home A & Home B & Home C & Mean change & 2 SE \\
    \midrule
    \multicolumn{6}{l}{\emph{Found within three rooms}} \\
    Last-seen (no language model) & 10 $\rightarrow$ 78 & 33 $\rightarrow$ 100 & 27 $\rightarrow$ 64 & $+56.9$ & 20.6 \\
    History and notes   & 40 $\rightarrow$ 78 & 50 $\rightarrow$ 100 & 82 $\rightarrow$ 82 & $+29.3$ & 30.1 \\
    Claim store         & 20 $\rightarrow$ 33 & 50 $\rightarrow$ 25  & 73 $\rightarrow$ 36 & $-16.0$ & 30.1 \\
    ACE-style playbook  & 10 $\rightarrow$ 33 & 83 $\rightarrow$ 25  & 55 $\rightarrow$ 55 & $-11.7$ & 48.6 \\
    \midrule
    \multicolumn{6}{l}{\emph{First room right}} \\
    Last-seen (no language model) & 10 $\rightarrow$ 56 & 33 $\rightarrow$ 50 & 18 $\rightarrow$ 55 & $+32.9$ & 17.0 \\
    History and notes   & 20 $\rightarrow$ 44 & 17 $\rightarrow$ 50 & 45 $\rightarrow$ 73 & $+28.4$ & 5.2 \\
    Claim store         & 10 $\rightarrow$ 11 &  0 $\rightarrow$ 0  &  9 $\rightarrow$ 27 & $+6.4$  & 11.8 \\
    ACE-style playbook  &  0 $\rightarrow$ 11 & 17 $\rightarrow$ 0  &  9 $\rightarrow$ 27 & $+4.2$  & 21.3 \\
    \midrule
    Questions, day 14 $\rightarrow$ 32 & 10 $\rightarrow$ 9 & 6 $\rightarrow$ 4 & 11 $\rightarrow$ 11 & & \\
    \bottomrule
  \end{tabular}
  \caption{\textbf{First day of each illness, day 14 $\rightarrow$ day 32}, on the 14 room-changing objects. Three households, 24 questions a day. Percentages; mean change and two SE from paired household differences. Measured against each method's own mean over the four preceding days, the first-room-right drops are: last-seen 59 then 31, history only 47 then 25, history and notes 57 then 29, claim store 51 then 48, ACE-style playbook 66 then 63. The paired changes in drop, with two SE, are $-27.4 \pm 14.4$ for last-seen, $-22.7 \pm 10.6$ for history only, $-28.1 \pm 6.6$ for history and notes, $-3.5 \pm 22.6$ for the claim store and $-2.9 \pm 27.0$ for the ACE-style playbook. Objects never found: last-seen 21 $\rightarrow$ 6, history and notes 11 $\rightarrow$ 4, claim store 14 $\rightarrow$ 16, ACE-style 15 $\rightarrow$ 14.}
  \label{tab:second}
\end{table}

\begin{table}[H]
  \centering
  \small
  \setlength{\tabcolsep}{4pt}
  \begin{tabular}{lccccc}
    \toprule
    Method & Home A & Home B & Home C & Mean change & 2 SE \\
    \midrule
    Last-seen (no language model) & 25 $\rightarrow$ 85 & 64 $\rightarrow$ 100 & 27 $\rightarrow$ 67 & $+45.1$ & 14.6 \\
    History and notes   & 50 $\rightarrow$ 85 & 73 $\rightarrow$ 100 & 82 $\rightarrow$ 83 & $+21.1$ & 20.1 \\
    Claim store         & 33 $\rightarrow$ 54 & 64 $\rightarrow$ 33  & 73 $\rightarrow$ 42 & $-13.6$ & 34.1 \\
    ACE-style playbook  & 25 $\rightarrow$ 54 & 91 $\rightarrow$ 50  & 55 $\rightarrow$ 58 & $-2.8$  & 40.8 \\
    \midrule
    Questions, day 14 $\rightarrow$ 32 & 12 $\rightarrow$ 13 & 11 $\rightarrow$ 6 & 11 $\rightarrow$ 12 & & \\
    \bottomrule
  \end{tabular}
  \caption{\textbf{Found within three rooms, day 14 $\rightarrow$ day 32, on all 22 moved objects of the first illness.} Eight of these do not change room in the second illness, which makes day 32 easier for every method. The pattern matches Table~\ref{tab:second}.}
  \label{tab:allmovers}
\end{table}

\begin{figure}[H]
    \centering
    \includegraphics[width=\linewidth]{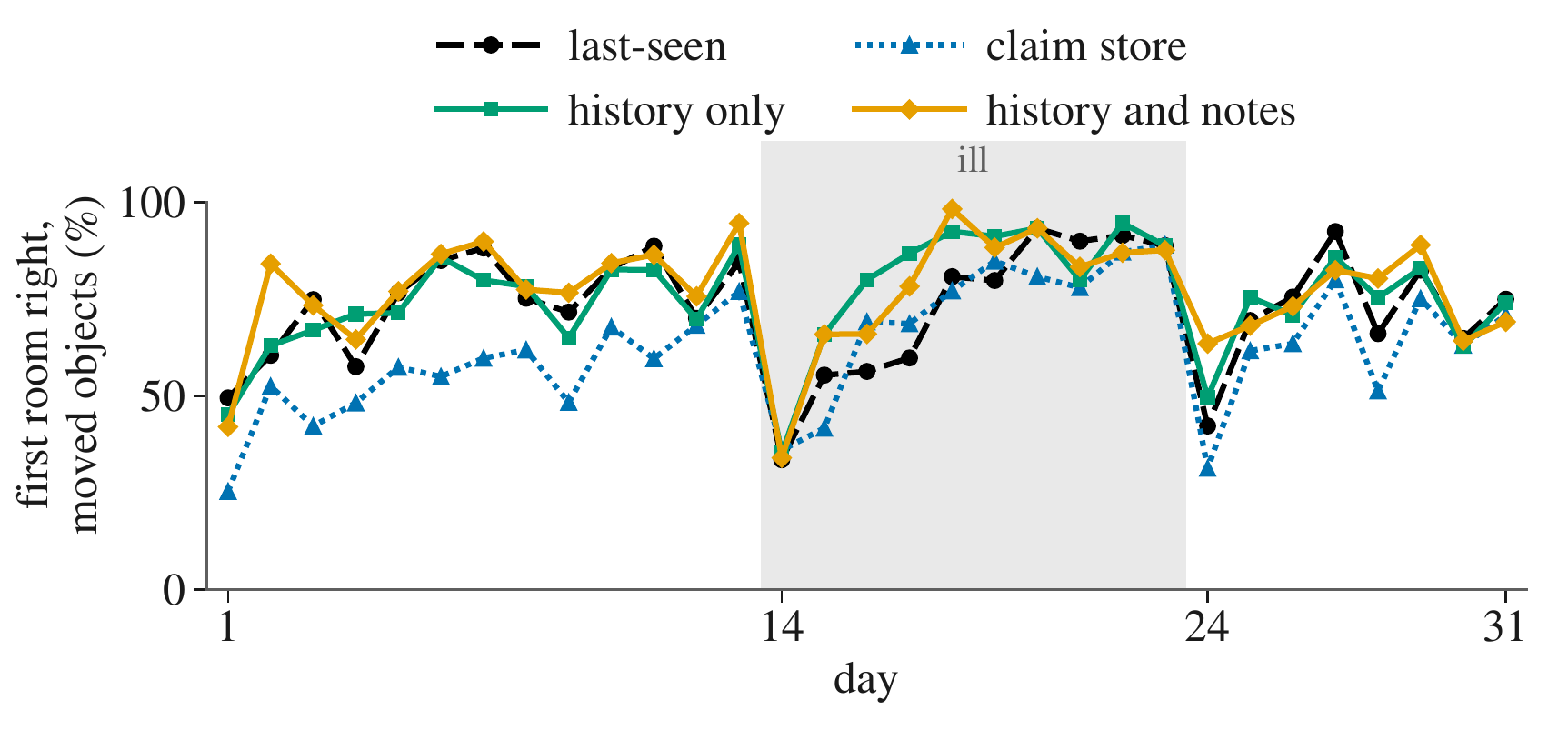}
    \caption{\textbf{The first illness, day by day.} First room right on the moved objects, ten households at 8 questions a day; sick days shaded. Every memory drops to about a third on day 14 and climbs back within the illness. The drop on day 24, when normal life returns, is similarly deep. The history-only line tracks history and notes through the whole illness, and both climb back faster than last-seen.}
    \label{fig:first}
\end{figure}

\section{Implementation Details}
\label{app:ace}

\paragraph{Two ACE-style variants.} The ten-household run (Table~\ref{tab:accuracy}, Figure~\ref{fig:first}) and the 50-day run (Figure~\ref{fig:second}, Table~\ref{tab:second}) use different ACE-style implementations. The ten-household variant reflects once a night, groups merge candidates by shared words, and merges by deterministic concatenation with no model involved. The 50-day variant reflects up to three times when the first room was wrong, ranks candidate pairs by embedding, hands at most four pairs a night to the model, and has the model judge whether they say one thing and write the merged text. Both share the differences from the released ACE code below.

\paragraph{Differences from ACE.} Our implementation keeps ACE's three-role division (a generator, a reflector and a curator) but differs from the released code in several ways. The released reflector loops up to three rounds and only when the answer was wrong; ours runs unconditionally each night. The released grow-and-refine step groups bullets by sentence embedding at cosine 0.90 and has the model author the merged text; our two variants handle this as described above. Pairs are restricted to notes with the same condition. The released code defines add, update, merge and delete but executes only add; ours runs rewrite, attach-evidence and join, and never deletes. The condition field on each note is our addition and is not part of ACE. The generator here is the robot's own room choices, and credit is whether the first room observed held the object.

\paragraph{The 50-day variant's merge step, as first run and as rerun.} In our first 50-day run, the merge call's schema typed the fields naming which note to keep and which to fold in as free strings. The model answered 356 of the 359 pairs it was shown, but 329 of those answers named something that was not a note (``both,'' ``none'') and were discarded without being counted. Only 7 notes were merged in two households over 100 nights. We changed the schema to an enumeration of the note ids in the prompt and reran on the same three households, seed and bank. Nothing is discarded now. Of 517 proposed pairs the model answered 424, merged 26 and rejected 396 as different notes. Table~\ref{tab:second}, Table~\ref{tab:allmovers} and Figure~\ref{fig:second} report the rerun. First room right is identical to the first run in every cell. Found within three rooms moved 11 points toward zero in Table~\ref{tab:second} and 23 in Table~\ref{tab:allmovers}. The first-day drops went from 53 and 55 to 66 and 63, with two-SE bands less than half as wide. Grow-and-refine does little on this task because the pairs an embedding ranks most alike are, in these households, notes about different objects, and the model says so when the schema lets it.

\paragraph{The history window.} The sighting history shows the newest forty sightings of the asked object; at 24 questions a day the cap binds on most questions. We did not run a larger window, so we cannot say the cap is free. The analyses we did run show no obvious association with reduced accuracy: history and notes' lead over last-seen is no smaller on questions where the cap bit, accuracy does not step down where the dated lines stop, and on questions the cap never touched the lead still falls from $+4.4$ at 4 questions a day to $+0.0$ at 24.

\paragraph{Room-choice prompt.} In the ten-household run the room-choice prompt named the room before its reason; a paired check found this does not move the day curves.

\paragraph{Cost.} A stored sighting is about 200 bytes. The robot stores 105 to 130 sightings a day at 8 questions a day and 310 to 370 at 24, depending on the method, which is under 75 KB a day. Prompt sizes are means over all calls, converted from characters at 3.3 to 3.4 characters per token. Per room choice, the claim store averages 3,499 prompt tokens on the ten households and 3,784 on the 50-day households, history only 2,932 and 3,500, and history and notes 5,320 and 6,120. Per nightly update, the claim store averages 9,467 and 23,953 prompt tokens and history and notes 8,119 and 20,039. On the ten households, the claim store writes about 1,950 tokens a night and history and notes about 1,350.

\section{Reading the Sighting History When Choosing a Room}
\label{app:log}

\paragraph{How the claim store and history and notes differ.} Both write notes in the same format, each with a condition and support counts, using the same nightly edits (add, revise, record evidence). Their prompts differ in four places. At room choice, history and notes also sees the asked object's sighting history, with a preamble naming the robot's three sources (its observations, its record and its notes) and a closing line asking what the residents are likely doing at that time of day. Its nightly prompt caps the notes at 16 and says the record is kept, so the notes need not copy it. Its system line asks the model to write down what it has worked out about the people in the home, where the claim store's asks for where things are kept. By day 14 in one household, the claim store held 61 notes, mostly inventory entries (``The kettle is kept on the kitchen counter''), and history and notes held 23, mostly about people and routines (``Tomas works from the office desk during the day and into the evening''). Because the two methods observe different rooms, they also see different things, so they are compared as whole methods.

\paragraph{Separating the history.} A variant of the claim store adds only the history block to its room-choice prompt and changes nothing else. On the ten households, a rerun of the claim store scores 75.8 first room right; showing it the day's own sightings raises this to 79.9 ($+4.1 \pm 2.8$), and showing it the full sighting history to 84.1 (a further $+4.2 \pm 3.0$), against 85.6 for history and notes. History only isolates the notes from the other side: it uses the room-choice prompt of history and notes with the notes removed.

\paragraph{Where the edge over last-seen comes from.} The language model's edge over last-seen does not come from the first room. On the 20 questions, out of 2,480, where neither method has yet seen the object in its new spot, both are right 15.9\% of the time. The edge comes after a wrong first room: on the first sick day, history and notes finds moved objects within three rooms 57\% of the time against last-seen's 24\%, and by days 15--17 it has seen the moved objects in their new spots on 95\% of questions against 59\% for last-seen. Seeing is not enough on its own: by the same days the claim store has seen the moved objects in their new spots on 96\% of questions and still gets the first room right on only 62\% of those questions, because it rewrote those sightings into summaries it cannot read back.

History only makes the same point from the other side. Across the ten households it trails history and notes by 0.4 and 0.7 points and leads the claim store by 8.2 and 9.0 (two SE 3.9 and 3.7, same sign in 9 and 10 of 10). On the day curve it tracks history and notes through the whole illness: 81.0 on days 10--13, 35.2 on day 14, 49.6 on day 24, against 85.3, 34.0 and 63.5 for history and notes and 81.6, 33.5 and 42.2 for last-seen.

How much history and notes improves over last-seen depends on how much the robot observes. On the five other households, history and notes leads last-seen by $+4.4$ and $+5.0$ points at 4 questions a day, in all five households, by $+3.5$ and $+2.6$ at 8, and by $+1.2$ and $+0.1$ at 24. The advantage is recovery from a wrong first room, so it grows when the robot gets fewer observations a day. With plenty of observations the sighting history alone is enough.

\section{The First Question About Each Object on the First Day of Each Illness}
\label{app:first}

We restricted day 14 and day 32 to the first question asked about each of the 14 room-changing objects, before that day's later searches can help. This leaves 13 object-days at day 14 and 11 at day 32 across the three households. On found within three rooms, last-seen goes from 23\% to 85\% and history and notes from 60\% to 92\%, while the claim store goes from 52\% to 53\% and the ACE-style playbook from 53\% to 60\%. A stricter version, keeping only objects that no earlier observation that day had already revealed, leaves 2 to 5 object-days per method and cannot support a comparison, because every observation records every object in the room.

\section{Deleting the First Illness From Memory}
\label{app:delete}

We replayed last-seen, history only and history and notes on the three 50-day households with every sighting dated days 14--23 dropped as it arrived. Bank, questions, seed, three-room budget and room-choice rule were unchanged. The robot still observed rooms and was scored on those days; it just carried nothing out of them. Sightings from day 24 on accumulated normally. A separate check, restricted to the first question about each object on days 14 and 32, gives the same recurrence picture without deletion (Appendix~\ref{app:first}). For history and notes, nights 0--13 made the same edits as the original run, days 1--13 chose the same rooms, and nights 14--23 made no update.

\paragraph{Last-seen.} On the 14 room-changing objects, day-32 first room right went from 56, 50 and 55 in the three households to 22, 0 and 45, and found within three rooms from 78, 100 and 64 to 22, 0 and 45. Measured against the days 28--31 mean, the day-32 drop went from 31.4 to 60.8 points on first room right (day 14: 58.8), a change of $+29.4 \pm 26.3$, and from 18.1 to 77.4 on found within three rooms (day 14: 76.5). In the original run, every one of last-seen's day-32 second- or third-room finds in a first-illness room came from a day-23 sighting.

\paragraph{History only.} Day-32 first room right went from 6 of 9, 2 of 4 and 8 of 11 questions to 4, 2 and 6, a change of $-13.5 \pm 13.7$ points. The day-32 drop went from 24.7 to 32.6 points ($+7.8 \pm 14.6$; day 14: 47.4). On found within three rooms the day-32 drop went from $-2.6$ to 26.2, rising in all three households.

\paragraph{History and notes.} Day-32 first room right went from 4 of 9, 2 of 4 and 8 of 11 questions to 3, 2 and 4, a change of $-15.8 \pm 21.5$ points. The day-32 drop went from 29.2 to 44.5 points ($+15.3 \pm 23.1$; day 14: 57.3), with most of the change in one household. On found within three rooms the day-32 drop went from 9.5 to 34.3, rising in all three households.

\paragraph{Last-seen updated nightly.} Without deletion, this variant drops 41.1 points on day 14 and 46.8 on day 32 ($+5.7 \pm 26.3$). It keeps every room the object has been seen in, as last-seen does, but adds the day's sightings only at night.

\section{What the Notes Held}
\label{app:conditions}

On night 31 of the ten-household run, the claim store, history and notes and the ACE-style playbook held 170, 142 and 114 notes that place an object where it went during the illness, 426 in all. Of these, 133 attach ``always,'' 238 a time of day, 12 a person being home and 43 some other condition. The twelve that name a person being home come from one household and one method and are true on well days too. None of the 426 mentions sickness.

\begin{figure}[H]
    \centering
    \includegraphics[width=\linewidth]{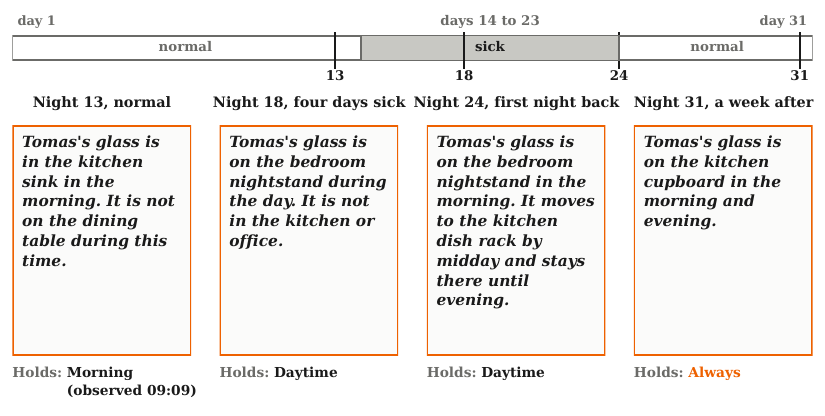}
    \caption{\textbf{One note across the month.} The claim store's note about the sick resident's glass, verbatim, on night 13 (normal), 18 (four days sick), 24 (first night back) and 31 (a week after). Each claim also speculates on when it holds true, but tends to write a time of day rather than the behavior of the individual. The note tracks the change and then overwrites it. By night 31 the bedroom spot is gone and the condition is ``always,'' on a spot that was true for ten days of the month.}
    \label{fig:life}
\end{figure}

\section{Telling the Model the Cause}
\label{app:told}

\begin{figure}[H]
    \centering
    \includegraphics[width=\linewidth]{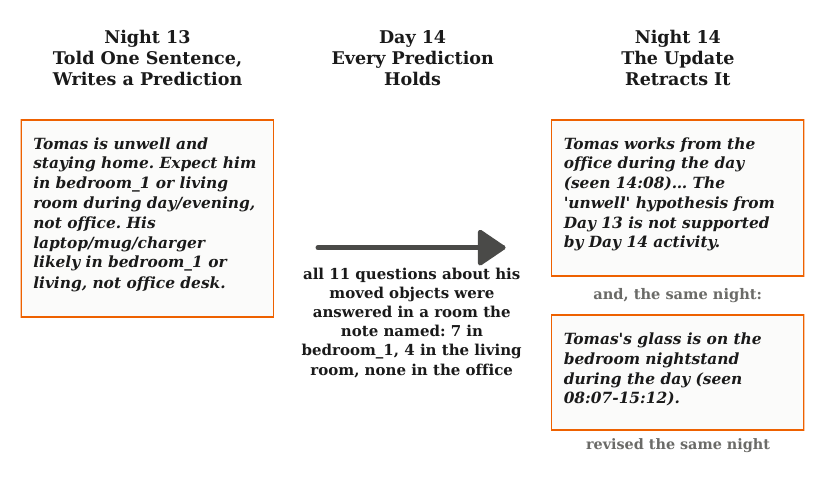}
    \caption{\textbf{The model predicts correctly when told the cause, and then retracts.} In this diagnostic run only, we tell the model on night 13 that a resident is sick and ask what will change. One household, history and notes. All 11 questions about the resident's moved objects the next day landed where the note said. The next nightly update dropped the hypothesis on one office sighting while, in the same pass, recording the evidence for it. Quotes verbatim; the ellipsis marks one omitted sentence. This is the strongest of the eight runs.}
    \label{fig:told}
\end{figure}

In eight runs we told history and notes one sentence on night 13, that a resident is sick and staying home, and asked what will change. The runs cover six households: two households were run twice, with different question banks. Every told run is history and notes; the cause goes to its nightly writer. All eight wrote a hypothesis naming the rooms the resident's objects would move to, before seeing a single changed sighting. On day 14, questions about those objects landed in a room the hypothesis named in 47 of 59 cases, from 3 of 6 in the worst run to 11 of 11 in the best. Figure~\ref{fig:told} shows the best case. The following night that run's model saw the resident in the office once and dropped the hypothesis. In the same pass it rewrote another note to record his glass on the bedroom nightstand from 08:07 to 15:12. In the normal routine that glass is never on the nightstand; through the illness it is there for 90\% of the working day. The model filed this sighting as a standing fact about the glass, a few lines below the note where it deleted the hypothesis.

Across the eight runs, three stopped conditioning on the illness within three nights, four more did so between nights 17 and 28, and one never did. Seven of eight filed a confirming sighting as an unconditioned standing fact in the same pass that weakened or removed the hypothesis. The nightly update sees that night's sightings and its own prior notes, but no earlier sightings. It could accumulate evidence through the notes it writes; here it mostly did not. We present this as a mechanism the nightly update can exhibit, with its rate. We have not shown that it causes the recurrence result.

\section{Limitations}
\label{app:limits}

The homes are simulated and perception on observing a room is perfect, so this isolates the memory problem from the perception problem. There is one kind of change and the second illness affects the same resident, so we test reuse of a recurring distribution rather than transfer to a new resident or a new kind of change. We use one language model with thinking disabled, and two summary designs. The recurrence result rests on three households, and its day-32 comparisons on 4 to 11 questions per household. Because the robot often sees a moved object in its new room within hours, the recurrence test only partly separates retention from relearning. Our ACE-style playbook is a variant of the published system, changed in the ways listed in Appendix~\ref{app:ace}, and its 50-day run merged notes only after a schema fix described there. We did not run a larger history window, and we did not test a fixed memory budget over months.

\end{document}